\documentclass[twoside]{article}

\usepackage[accepted]{aistats2027}

\usepackage[round]{natbib}
\renewcommand{\bibname}{References}
\renewcommand{\bibsection}{\subsubsection*{\bibname}}

\usepackage{times}

\usepackage{hyperref}
\usepackage{url}

\usepackage{amsthm}
\newtheorem{example}{Example}

\newtheorem{proposition}{Proposition}

\usepackage{mathtools} %
\usepackage{booktabs} %
\usepackage{tikz} %
\usepackage{researchpack}
\usepackage[capitalise]{cleveref}
\usepackage{wrapfig}
\usepackage{pifont}
\usepackage{enumitem}
\usepackage{algorithm}
\usepackage{algpseudocode}

\newcommand{\pal}{\texttt{PAL}\xspace}
\newcommand{\gasp}{\texttt{GASP!}\xspace}

\newcommand{\gaspOne}{\texttt{GASP!\kern-0.3em!}\xspace}
\newcommand{\gaspTwo}{\texttt{GASP!\kern-0.3em!\kern-0.3em!}\xspace}
\newcommand{\gaspThree}{\texttt{GASP!\kern-0.3emV}\xspace}

\newcommand{\lra}{\mathcal{LRA}}
\newcommand{\smtlra}{%
  \ifmmode
    \text{SMT}(\lra)%
  \else
    SMT(\ensuremath{\lra})\xspace%
  \fi
}

\hypersetup{
colorlinks=true,linkcolor=purple,citecolor=teal
}

\newcommand{\formula}{\ensuremath{\Phi}\xspace}
\newcommand{\formulaEx}{\ensuremath{\Delta}\xspace}

\newcommand{\pcInputFun}[1]{f_{#1}}
\newcommand{\pcProd}{\mathrm{Prod}}
\newcommand{\pcSum}{\mathrm{Sum}}

\newcommand{\pcPal}{\texttt{CiPAL}\xspace}

\newcommand{\Ygroup}[1][]{%
    \ensuremath{\vY_{\mathrm{group}_{#1}}}%
}

\usepackage[dvipsnames]{xcolor}
\usepackage{xspace} %
\newcommand{\av}[1]{\textbf{\textcolor{orange}{{av: #1}}}}
\newcommand{\lk}[1]{\textbf{\textcolor{ForestGreen}{{lk: #1}}}}
\renewcommand{\av}[1]{}
\renewcommand{\lk}[1]{}

\renewcommand{\paragraph}[1]{\textbf{#1}}

\begin{document}

\twocolumn[

\aistatstitle{\pcPal: Scaling Probabilistic Learning Under Algebraic Constraints}

\aistatsauthor{ Leander Kurscheidt \And Antonio Vergari}

\aistatsaddress{ University of Edinburgh \And  University of Edinburgh} ]

\begin{abstract}
Machine learning models often treat predictions as unconstrained points, although real outputs must satisfy semantic constraints: trajectories should avoid obstacles, physical quantities obey laws, and engineering designs must be feasible. 
To represent predictive uncertainty while respecting constraints, probabilistic algebraic layers \citep{DBLP:conf/uai/KurscheidtMSPV25} compositionally combines unconstrained probabilistic predictions with domain-specific algebraic constraints, restricting their support to outputs that satisfy these constraints.
However, exact normalization becomes inefficient as dimensionality grows. 
To overcome this limitation, we design a new family of plug-and-play differentiable prediction layers, \pcPal, that exploit structure common in real-world constrained problems: constraints often involve only subsets of variables; their feasibility boundaries are usually spatially localized;
as well as data, requiring to deal with constraints in an adaptive way.
We showcase \pcPal on real-world applications in trajectory prediction
and optical engineering, 
where it outperforms existing methods in both log-likelihood and point-prediction metrics, while guaranteeing the satisfaction of the constraints.
\end{abstract}

\section{Introduction}\label{sec:intro}

\begin{figure}[h]
    {
    \small
    \setlength{\fboxsep}{0pt}
    \begin{minipage}{0.28\linewidth}
    \centering
    Prediction:\\
    $p(\vY|\vx)$\\[1pt]
    \includegraphics[
        width=\linewidth,
    ]{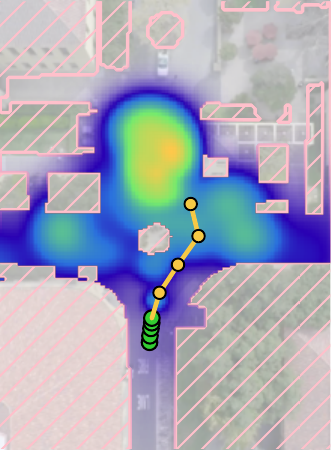}
    \end{minipage}
    \begin{minipage}{0.28\linewidth}
    \centering
    Query: Goal Area\\
    $p(\vY|Y_4{\in}\raisebox{-0.2em}{\includegraphics[height=1.1em]{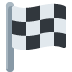}}, \vx)$\\[.3pt]
    \includegraphics[
        width=\linewidth,
    ]{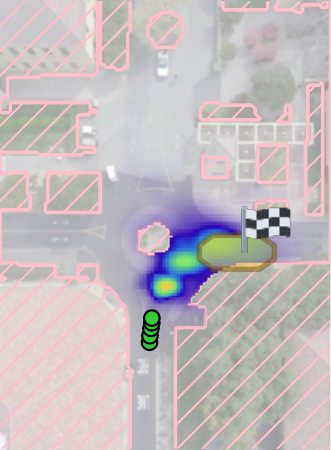}
    \end{minipage}
    \hfill
    \begin{minipage}{.42\linewidth}
    \centering
    Query: Collision Risk\\
    $\Pr(
    \vY{\in}\raisebox{-0.2em}{\includegraphics[height=1.0em]{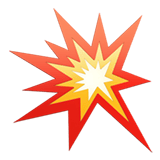}}
    |\vx){=}43\%$\\[.5pt]
    {
    \includegraphics[
        width=0.48\linewidth,
    ]{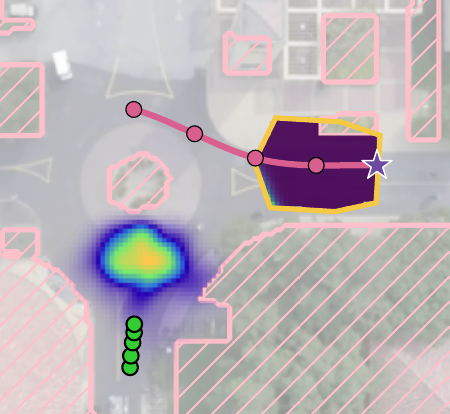}
    \includegraphics[
        width=0.48\linewidth,
    ]{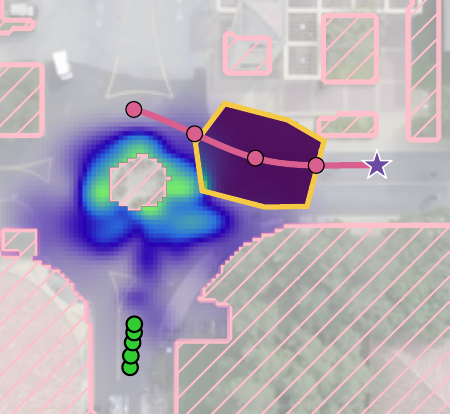}\\[1.8pt]
    \includegraphics[
        width=0.48\linewidth,
    ]{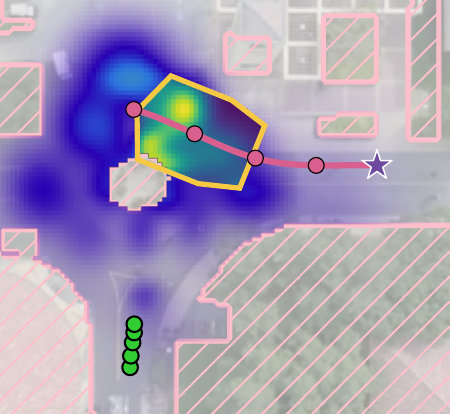}
    \includegraphics[
        width=0.48\linewidth,
    ]{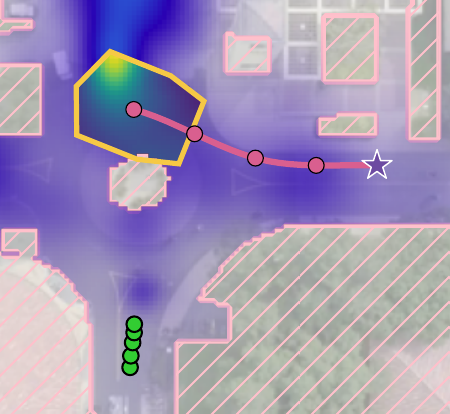}\\
    }
    \end{minipage}
    }
    \caption{\textbf{\pcPal enables exact semantic queries over constrained distributions} such as  predicting the future trajectory of another agent (left). We can then condition the predicted distribution on semantic events, such as reaching a chosen goal region (middle), or exactly compute structured event probabilities such as the risk of collision (right). 
    Details of the above settings in \cref{sec:exps}.
    }
    \label{fig:sdd_semantic_queries}
\end{figure}

Many real-world applications require the satisfaction of algebraic constraints: trajectories should avoid obstacles \citep{DBLP:conf/eccv/RobicquetSAS16, DBLP:conf/uai/KurscheidtMSPV25}, light polarization is physically bounded by its intensity \citep{redding2013compact}, and control systems must remain within polyhedral safety regions \citep{pmlr-v168-pfrommer22a}.
Therefore, probabilistic machine learning (ML) models should not only capture uncertainty over plausible outcomes, but also assign zero probability to outcomes that violate such constraints. 
In safety-critical applications, this provides a hard guarantee that invalid predictions are never sampled nor can be argmaxed \citep{DBLP:conf/aaai/GrivasVL24,DBLP:conf/iclr/StoianDCLG24,DBLP:conf/iclr/StoianG25}. 
When this does not happen, probabilities leaking over infeasible regions can severely underestimate the probability of feasible regions, leading to underestimate potential risks. 
This can be problematic, for example, when the risk of an action is the probability of a collision, see \cref{fig:sdd_semantic_queries} and \citet{9805817}.

Even in non-safety-critical applications, 
imposing constraints on ML models can yield substantial accuracy gains \citep{HANSEN2024133952}. 
This can be especially beneficial in long-running simulations, where even small constraint violations may accumulate over time and destabilize the simulation, as in climate or physical-system modeling \citep{beucler2019achieving, beucler2021enforcing}.
One way to achieve this is to design a model constrained by design, typically through a handcrafted reparameterization that maps an unconstrained space to the feasible region \citep{Zhou_2019_CVPR, RichterPowell2022NeuralCL} or by projecting points in the former to the latter \citep{DBLP:conf/iclr/StoianDCLG24,DBLP:conf/iclr/StoianG25}.
However, they can distort the underlying density and might not provide a principled treatment of uncertainty.

Probabilistic algebraic layers \citep[\pal]{DBLP:conf/uai/KurscheidtMSPV25} instead take a more probabilistic approach: they multiply an unconstrained density with  domain-specific algebraic constraints represented in satisfiability modulo theory \citep[SMT]{barrett2021satisfiability}. 
This restricts the distribution to feasible outputs while still allowing exact constrained log-likelihood evaluation when the distribution is modelled by piecewise polynomials \citep{DBLP:journals/corr/abs-2003-00126, DBLP:journals/corr/abs-1909-09362}. 
However, as exactly normalizing the resulting distribution, also known as computing the \textit{weighted model integral} (WMI) \citep{belle2015probabilistic,
morettin2019advanced}  scales poorly in high dimensions \citep{DBLP:journals/corr/abs-2003-00126, baldoni2011integrate}, \pal can efficiently deal only with lower-dimensional predictive distributions, as discussed in  
\cref{sec:background}.

In this paper, we establish that this point of view is overly pessimistic, as many real world problems exhibit structure that can be exploited to scale inference. 
To this end, we design a new family of plug-and-play differentiable prediction layers, \pcPal, that exploit structure common in real-world constrained problems:
\begin{enumerate}[topsep=0pt, itemsep=0pt, parsep=0pt, partopsep=0pt]
    \item[{\hyperref[sec:gasp_s1]{\textbf{S1}}}] \textbf{\textit{constraints decomposing}}   
    into non-overlapping sets of variables, possibly shared, for example spatial constraints over a time axis (\cref{sec:gasp_s1,sec:pcp_pal}).
    \item[{\hyperref[sec:gasp_s2]{\textbf{S2}}}] \textbf{\textit{sparseness of feasibility-boundaries}} when 
    using piecewise polynomials for densities, boundaries of the feasible set are usually spatially localized and do not cut every piece. By doing expensive constrained integration only where constraints actually matter, we can significantly speed up inference (\cref{sec:gasp_s2}).
    \item[{\hyperref[sec:gasp_s3]{\textbf{S3}}}] \textbf{\textit{adaptive grid resolution}} to capture predictions that are not distributed uniformly across the high-dimensional space, but concentrate in smaller regions. By increasing resolution in these regions, we can preserve high density fidelity in high dimensions (\cref{sec:gasp_s3}).
\end{enumerate}

\paragraph{Contributions.} By exploiting S1-3, we combine expressive probability distributions in high-dimensions with guaranteed constraint-satisfaction, exact log-likelihoods and probabilistic queries. 
In addition to this, we perform extensive experiments (\cref{sec:exps}) in which \pcPal outperform purely neural models in two real-world constrained prediction tasks in trajectory prediction and optical engineering.

\section{Constrained Probabilistic ML}
\label{sec:background}

\paragraph{Notation.}
We use uppercase letters to denote random variables (e.g., $X, Y$) and lowercase letters for their realizations (e.g., $x, y$). Bold symbols represent (random) vectors (e.g., $\vX, \vY$) and their joint assignments (e.g., $\vx, \vy$). Greek letters such as $\formulaEx$ or $\Phi$ denote logical formulas mapping real-valued variables to Boolean values (false or true). An assignment $\vx$ is said to satisfy a constraint $\formula$, written as $\vx \models \formula$, if substituting $\vx$ into $\formula$ evaluates it to true. The corresponding indicator function $\Ind{\vx \models \formula}$ equals 1 if $\vx$ satisfies $\formula$, and 0 otherwise.

\begin{figure}[!t]
    \centering
    {
\setlength{\fboxsep}{0pt}

\begin{minipage}[t]{0.28\linewidth}
    \centering
    $\left.
    \begin{array}{c}
        \includegraphics[width=0.8\linewidth]
        {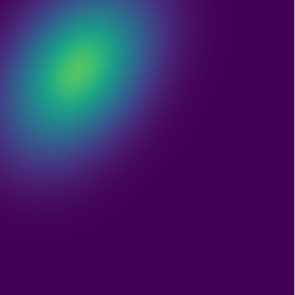}
        \\[2pt]
        \includegraphics[width=0.8\linewidth]
        {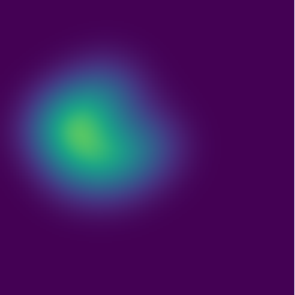}
    \end{array}
    \right\}$
\end{minipage}
\hfill
\begin{minipage}[t]{0.29\linewidth}
    \centering
    \vspace{-30pt}
    \raisebox{20pt}{\Large $\times$}\hspace{5pt}\fbox{%
        \includegraphics[width=0.75\linewidth]
        {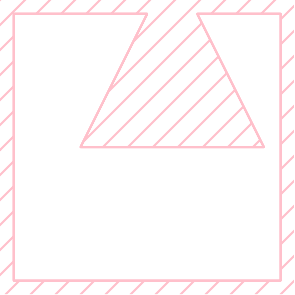}%
    }
\end{minipage}
\hfill
\begin{minipage}[t]{0.03\linewidth}
    \centering
    \textbf{\shortstack{\strut\\\strut}}\\[-10pt]
    \raisebox{40pt}{\Large $=$}
\end{minipage}
\hfill
\begin{minipage}[t]{0.28\linewidth}
    \centering
    $\begin{array}{c}
        \fbox{%
            \includegraphics[width=0.8\linewidth]
            {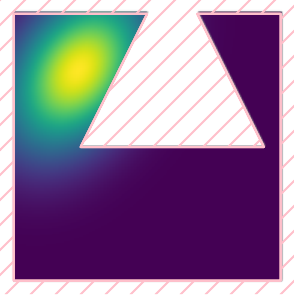}%
        }
        \\[2pt]
        \fbox{%
            \includegraphics[width=0.8\linewidth]
            {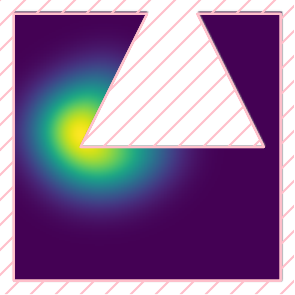}%
        }
    \end{array}$
\end{minipage}
}

    \caption{\textbf{Density$\times$constraints$\ =\ $constrained density:} \pcPal restricts an unconstrained density (here represented as marginals for two time-steps) to the feasible region using an SMT formula, which is \textit{shared} across time steps.}
    \label{fig:constraints_example}
\end{figure}

\begin{figure}[!t]
    {
    \setlength{\fboxsep}{0pt}
    \begin{minipage}{0.28\linewidth}
        \centering
        \textbf{Setting}\\[1pt]
        \fbox{%
            \includegraphics[
                width=\linewidth,
                trim={0 10 0 10},
                clip
            ]{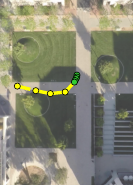}%
        }
    \end{minipage}
    \hfill
    \begin{minipage}{0.28\linewidth}
    \centering
    \textbf{NN+GMM}\\[1pt]
    \fbox{%
    \includegraphics[
        width=\linewidth,
        trim={0 10 0 10},
        clip
    ]{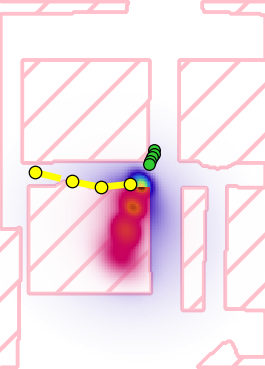}%
    }
    \end{minipage}
    \hfill
    \begin{minipage}{0.28\linewidth}
    \centering
    \textbf{NN+PcPAL}\\[1pt]
    \fbox{%
    \includegraphics[
        width=\linewidth,
        trim={0 10 0 10},
        clip
    ]{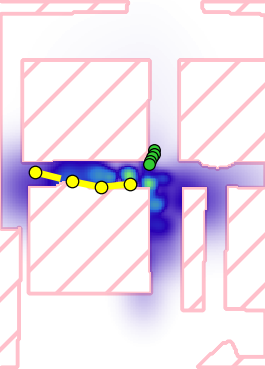}%
    }
    \end{minipage}
    }
    \caption{\textbf{The setting of the Stanford drone dataset.} One the left, one can see the green observed trajectory with its ground-truth future trajectory in yellow. In the middle an GMM(RNN) prediction and on the right \pcPal.}
    \label{fig:sdd_overview}
\end{figure}

\paragraph{\smtlra Constraints.}
To model domain constraints, we use \smtlra, the SMT fragment over linear real arithmetic (\citealt{barrett2021satisfiability}). 
\smtlra represents constraints using piecewise-linear boundaries over both real-valued and boolean variables. 
We consider quantifier-free SMT formulas over continuous variables $\vX$ \citep{DBLP:journals/corr/abs-2003-00126}, consisting of linear (in)equalities of the form $\sum_i a_i X_i \bowtie b$, where $\bowtie \in \{\le, =\}$, combined using Boolean connectives for conjunctions ($\wedge$), disjunctions ($\vee$), and negation ($\neg$). 
{%
\setlength{\topsep}{3pt}
\setlength{\partopsep}{0pt}
\setlength{\parskip}{0pt}
\begin{example}[SMT formula]
    Consider the SMT formula $ \formulaEx(\vX)=\bigwedge_{t=1}^n \formulaEx_t(\vX^{(t)})$ over the sequence of bivariate variables $\vX^{(1)}=\{X^{(1)}_1, X^{(1)}_2\}, \vX^{(2)},\dots, \vX^{(n)}=\{X^{(n)}_1, X^{(n)}_2\}$, which defines a polytope-constraint per time-step $t$:
    \vspace{-2mm}
    \begin{equation*}
    \begin{aligned}
    \formulaEx_t(\vX^{(t)})&=X^{(t)}_1{\in}[0,2]\ {\wedge}\ X^{(t)}_1{\in}[0,2]\\
    &\phantom{=}{\wedge}(X_2{\leq}1{\ \vee\ }X_2{>}2X_1{\ \vee\ }X_2{>}4.75{-}2X_1).
    \end{aligned}
    \end{equation*}
 This formula represents the non-convex polytope illustrated in \cref{fig:constraints_example} for $t=1$.
\label{ex:smt}
\end{example}
}%
More generally, \smtlra can represent real-world, and disconnected feasibility constraints, such as the walkable area in \cref{fig:sdd_semantic_queries,fig:sdd_overview}.

\paragraph{PAL.}
Our task is to learn a parameterized conditional distribution of $\vY$ given $\vX$, i.e. $p_{\boldsymbol{\Theta}}(\vY \mid \vX)$ from a dataset $\mathcal{D}=(\vx^{(i)},\vy^{(i)})$ consisting of realizations of the continuous random variables $\vX$ and $\vY$.
For example, $\vX$ can be the observed point in a trajectory and $\vY$ the continuation to predict (\cref{fig:sdd_semantic_queries}).
In contrast to the classical unconstrained setting, the support of $\vY$ is restricted by a formula $\formula \in \smtlra$, ensuring that $p_{\boldsymbol{\Theta}}(\vY{\models}\neg \formula \mid \cdot){=}0$. 
\pal
tackles the problem as follows: it uses an unconstrained piecewise polynomial density $q$ over $\vY$   with parameters $\boldsymbol{\lambda}$, which are predicted by a neural network $f_{\boldsymbol{\psi}}$. Then,  $q$ is restricted to the feasible set and renormalized as:
\begin{equation}
    p_{\boldsymbol{\Theta}}(\vY \mid \vx) = \frac{q(\vY; \boldsymbol{\lambda}{=}f_{\boldsymbol{\psi}}(\vx)) \Ind{\vY \models \phi}}{\int q(\vy'; \boldsymbol{\lambda}{=}f_{\boldsymbol{\psi}}(\vx)) \Ind{\vy' \models \phi} d\vy'}
    \tag{PAL}
    \label{eq:pal}
\end{equation}
It is important to notice that the normalizing factor is not a constant but a function of $\vx$, or, if we change perspective, a function of $\boldsymbol{\lambda}$. 
Efficiently computing the denominator, the WMI \citep{morettin2019advanced}, is a major challenge. \pal approaches the problem by considering the symbolic computation in $\boldsymbol{\lambda}$, where $\boldsymbol{\lambda}$ is a simple parametrization via the coefficients of the monomials:
\begin{align}
    I(\boldsymbol{\lambda})&=\int q(\vy'; \boldsymbol{\lambda}) \Ind{\vy' \models \phi} d\vy'
    \tag{PAL WMI}\label{eq:pal_wmi}\\
    &=\int \sum_\alpha \lambda_\alpha \prod_i ({y'_i})^{\alpha_i} \Ind{\vy' \models \phi} d\vy'\\
    &= \sum_\alpha \lambda_\alpha \underbrace{\int\prod_i ({y'_i})^{\alpha_i} \Ind{\vy' \models \phi} d\vy'}_{:=M_{\alpha}(\phi)\text{, independent of $\boldsymbol{\lambda}$}}
    \label{eq:pal_wmi_mon}
\end{align}
So by precomputing, or ``compiling'', the constrained moments $M_{\alpha}(\phi)$ only \textit{once}, 
\ref{eq:pal} turn into an efficient, exact and differentiable operation. 
Computing $M_{\alpha}(\phi)$ is done through \gasp, which is a GPU-accelerated integration-backend that works through enumerating the feasible area and numerically integrating the polynomial over it without approximation. Unfortunately, computing even a single constrained moment $M_\alpha(\phi)$ is NP-hard \citep{baldoni2011integrate} and becomes prohibitively expensive as dimensionality grows. At the same time, a polynomial of total degree $m$ in $n$ dimensions requires $\binom{n+m}{m}$ such moments. Together, these effects limit \pal to lower-dimensional predictions.

\section{Exploiting Structure}
\begin{figure}[h]
    \begin{center}
    \begin{minipage}[t]{0.4\linewidth}
    \hspace{25pt}\textbf{Dimensionality}\\[1pt]
    \includegraphics[
        width=\linewidth,
        clip
    ]{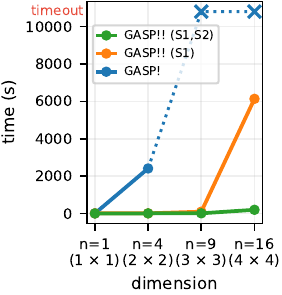}
    \end{minipage}
    \hspace{10pt}
    \begin{minipage}[t]{0.38\linewidth}
    \hspace{25pt}\textbf{Resolution}\\[1pt]
    \includegraphics[
        width=\linewidth,
        clip
    ]{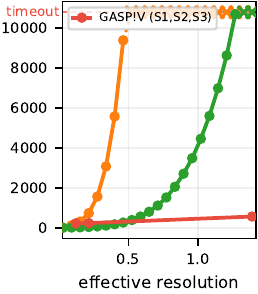}
    \end{minipage}
    \end{center}
    \caption{\textbf{Exploiting structure scales \gasp in both dimensionality and resolution.} Decomposition (\hyperref[sec:gasp_s1]{\textbf{S1}}) and sparsity (\hyperref[sec:gasp_s2]{\textbf{S2}}) enable high-dimensional integration (left), while adaptive refinement (\hyperref[sec:gasp_s3]{\textbf{S3}}) provides high resolution where needed (right). Details in \cref{app:s2_sparsity_plot}.}
    \label{fig:gasp_structures_overview}
\end{figure}
In this section we scale \pal to higher-dimensional problems by exploiting structure commonly found in many real-world applications. 
We focus on three complementary structures that,
as the WMI grows intractable in higher dimensions, try to decouple the integral complexity from the dimensionality. 
We solve this by focusing on problems with constraint decomposition (\hyperref[sec:gasp_s1]{\textbf{S1}}), which appears in a range of applications like obstacles in trajectory prediction \citep{DBLP:conf/uai/KurscheidtMSPV25} or modeling fluids \citep{ohana2024well}. 
Secondly (\hyperref[sec:gasp_s2]{\textbf{S2}}), when enumerating the pieces of the piecewise function for the integral, we exploit the fact that usually, the share of cells, i.e., the pieces of the domain of the polynomials that form a \textit{grid}, that are neither fully inside or outside the SMT constraint  shrinks with higher dimensionality or higher grid resolutions. 
And third (\hyperref[sec:gasp_s3]{\textbf{S3}}), we adapt the resolution of the grid to the parts of the space that matter, as usually predictions are not uniformly distributed but cluster in specific local regions.
These computational benefits are summarized in \cref{fig:gasp_structures_overview}.

\subsection{S1: Scaling up \gasp}
\label{sec:gasp_s1}
\begin{figure}[h]
    \begin{minipage}[t]{0.28\linewidth}
    \hspace{30pt}\textbf{Time}\\[1pt]
    \includegraphics[
        width=\linewidth,
        clip
    ]{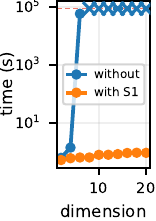}
    \end{minipage}%
    \hfill
    \begin{minipage}[t]{0.29\linewidth}
    \hspace{30pt}\textbf{Space}\\[1pt]
    \includegraphics[
        width=\linewidth,
        clip
    ]{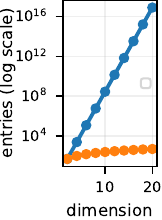}
    \end{minipage}%
    \hfill
    \begin{minipage}[t]{0.4\linewidth}
    \vspace{-18pt}
    \caption{\textbf{Constraint decomposition (S1) scales \gasp in both time and space (number of basis functions).} Without decomposition, both grow exponentially with dimensionality. See \cref{app:s1_gasp_benchmark}.}
    \label{fig:scaling_up_gasp}
    \end{minipage}
\end{figure}

At the core of \pal is the exact computation of integrals over the constrained domain, which boil down to compute smaller integrals over many convex polytopes \citep{morettin2019advanced,DBLP:conf/uai/KurscheidtMSPV25}. 
Figure~\ref{fig:scaling_up_gasp} illustrates the two sides of this problem: computing all required moments becomes increasingly expensive (left), while the number of moments that need to be represented grows rapidly (right).
To overcome this dimensionality bottleneck, we exploit that real-world constraints like obstacles in multi-step trajectory prediction (\cref{fig:sdd_semantic_queries}) often decompose into independent, lower-dimensional constraints (S1). 
That is, they can be written as 
\begin{equation}
\formula(\vY)
= \bigwedge\nolimits_i \formula_i(\Ygroup[i]),
\label{eq:decomp}
\end{equation}
where the variable scopes $\mathrm{group}_i$ are pairwise disjoint and the dimensionality of each group is bounded, or $\max_i |\Ygroup[i]| \leq d$. \cref{ex:smt} satisfies this desiderata, as well as the trajectories from the Stanford drone dataset \citep{DBLP:conf/uai/KurscheidtMSPV25} and light polarization in optical engineering \citep{redding2013compact}.

Our first question is whether this structure is sufficient to make the computation of the WMI underlying \pal (\cref{eq:pal_wmi}) efficient in high dimensions. 
In general, integrating even a bounded-degree polynomial over a single polytope is intractable when the dimensionality is allowed to grow \citep{baldoni2011integrate}. 
If the constraints decompose into bounded-dimensional groups as in \cref{eq:decomp}, and the polynomial $q$ decomposes over the same groups through sums and products, then the WMI can be computed tractably in $|\vY|$ under mild assumptions on the group-wise constraints (see \cref{app:tractability_wmi}). Thus, decomposability provides a tractable route to the integrals required by \pal, but requires the predictive density under the integral to factorize according to the constraint structure, which might hinder estimating uncertainty, as it would rule out dependencies across constraint groups and immensely limit the expressivity of our predictive distribution.

\subsection{\pcPal: Exploiting Factorization without Restricting Distributional Dependence}
\label{sec:pcp_pal}
To scale \pal without sacrificing these dependencies, we therefore need to decouple the factorization used to compute the constrained integral from the dependence structure of the predictive distribution.
We do this by leveraging tractable probabilistic models named \textit{circuits}. 

\paragraph{Probabilistic circuits}
\citep[PCs]{darwiche2003differential,vergari2019tractable,
ProbCirc20}  construct expressive distributions by composing simple, typically univariate, distributions into hierarchical mixtures by interleaving sum and product operations.
Furthermore, they allow probabilistic operations such as integration to be decomposed along the induced variable hierarchy \citep{DBLP:conf/nips/VergariCLTB21}.
This allows us to represent dependencies across constraint groups while breaking down the integral over smaller integrals over each variable group, if the PC structure is built to reflect the constraint structure of \cref{eq:decomp}. %

In this work, we use a simple class of tensorized PCs \citep{loconte2025relationship}. We consider univariate distributions as base input distributions. Each variable $X_i$ is mapped into a vector $\pcInputFun{X_i}$ via $[\pcInputFun{X_i}(X_i)]_j=p_j(X_i)$, %
where $p_j$ is a density of a univariate distribution that supports tractable computation of the CDF.
Rather than concatenating these representations, we treat them as separate vectors throughout the circuit. Each vector $\vl$ is associated with a scope $\mathrm{sc}(\vl)$, indicating the subset of variables on which its computation depends.
Acting on these vectors, we use two types of layers \citep{loconte2025relationship}, although more are feasible: the Product-layer ($\pcProd$), which combines two layers using an element-wise product, and  the sum-layer ($\pcSum$) to transform a vector linearly:%
{%
\[
\begin{aligned}
\pcProd &: (\mathbb{R}^a,\mathbb{R}^a) \rightarrow \mathbb{R}^{a}
& \pcProd(\vl, \vl') &:= \vl \odot \vl'\\%
\pcSum &: \mathbb{R}^a \rightarrow \mathbb{R}^o
& \pcSum(\vl) &:= \vW \vl \\%
\end{aligned}
\]
}%
Here, we assume that the layers $\vl$ and $\vl'$ are defined over disjoint sets of variables, i.e., the PC is \textit{decomposable}.
We can combine these layers until we reach a scalar which is the output of the PC.
Computing the normalizing constant of an unconstrained density, \textit{or one restricted by independent bounds on each variable}, over such a circuit is now straightforward:
Multilinearity of $\pcProd$ and $\pcSum$ enables us to push the actual integral from the overall computation back to the input-functions.
For example, for $\pcProd(\vl', \vl'')$ with $l_i':=p_i'(x_1)$ and $l_i'':=p_i''(x_2)$ we have $\int_{A\times B}\pcProd(\vl', \vl'')d\vx=\pcProd(\int_{A}\vl', \int_{B}\vl'')$ with vectors $(\int_A \vl')_i:=\int_A p_i'(x_1)dx_1$ and $(\int_B \vl'')_i:=\int_B p_i''(x_2)dx_2$. 

\paragraph{\pcPal.} By pushing down the WMI integral along the multi-linear transformations, we split it into lower-dimensional WMI integrals over both $\mathrm{sc}(\vl)$ and $\mathrm{sc}(\vl')$, as long as the constraint decomposes accordingly to the scopes. 
Using this structure, we can therefore exploit the factorization of the constraints for tractable integration without imposing the same factorization on the predictive distribution. 
We call this construction \pcPal. For polynomial input distributions, each of these lower-dimensional sub-circuits is itself polynomial. We can therefore apply the \pal machinery \textit{locally}: We can pre-compute the constrained moments $M_\alpha(\formula_i)$ over only the sub-circuits, which still encode polynomial distributions, but of limited dimensionality, and use them to efficiently and exactly evaluate the integral during training (see \cref{app:pc_pal} for more details).
In practice, we use splines as PC inputs and use them to interpolate an unconstrained probabilistic model (see \cref{sec:exps} and \cref{sec:exp_sdd_pcpal}).

\subsection{S2: Exploiting Sparseness}
\label{sec:gasp_s2}
\begin{figure}[!t]
    {
    \setlength{\fboxsep}{0pt}
    \begin{minipage}{0.28\linewidth}
    \centering
    \small $4\times6$ pieces\\[1pt]
    \includegraphics[
        width=\linewidth,
        clip
    ]{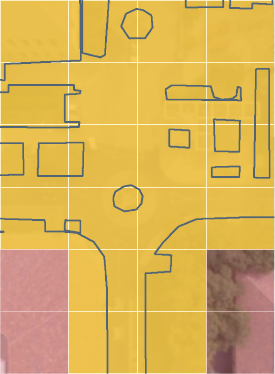}
    \end{minipage}
    \hfill
    \begin{minipage}{0.28\linewidth}
    \centering
    \small$9\times11$ pieces\\[1pt]
    \includegraphics[
        width=\linewidth,
        clip
    ]{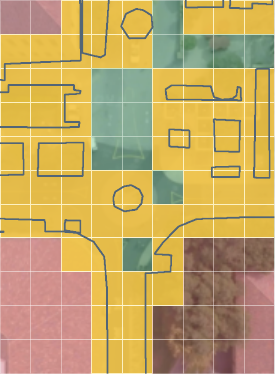}
    \end{minipage}
    \hfill
    \begin{minipage}{0.28\linewidth}
    \centering
    \small$17\times23$ pieces\\[1pt]
    \includegraphics[
        width=\linewidth,
        clip
    ]{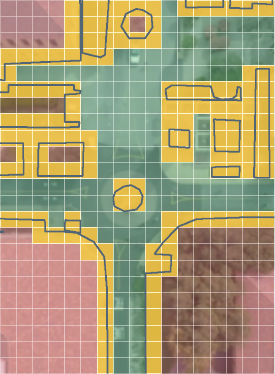}
    \end{minipage}
    }
    \caption{\textbf{\pcPal: exploiting sparsity lets us focus on cut cells for the full compilation}. We show the increasing sparsity of the cells intersecting with constraints with higher resolution (red: outside, yellow: cut, green: inside). The full WMI is only needed for cut cells, an effect that becomes even more pronounced in higher dimensions.}
    \label{fig:sparsity}
\end{figure}

In Section~\ref{sec:pcp_pal}, we showed how constraint decomposition allows \pcPal to scale to high-dimensional problems while keeping the dimensionality $\dim(\Ygroup[i])\eqqcolon n_{i}$ of individual constraint groups $\Ygroup[i]$ bounded (S1). We now turn to the complementary problem of making integration within each group $\Ygroup$ more efficient, allowing \pcPal to handle larger groups. To this end, we exploit the spatial locality of feasibility boundaries (S2). Since \pcPal uses piecewise densities, their pieces induce a partition or grid of each $\Ygroup$ into $n_i$-dimensional cells. 
In order to compile our constrained moments, we first have to integrate over each cell in such a grid.

To speed this up, we first realize that the constrained moment computation is only really needed when we have a feasibility boundary cutting a cell. Otherwise we are either completely infeasible (no integration needed) or completely feasible. For entirely feasible cells, no constrained integration is necessary: using the PC structure of \pcPal, we can reduce the integral over the full cell to a combination of simple one-dimensional interval integrals over the input distributions \citep{ProbCirc20}, as cells are axis-aligned. See \cref{app:s2_sparse_algo} for pseudo-code.

Fortunately, for most real-world problems the boundaries do not lie dense in the space. For the feasibility boundary to intersect every cell, it would have to repeatedly weave back and forth through the space at the scale of the partition, constantly alternating between feasible and infeasible regions. Such a highly oscillatory boundary is unlikely for the structured geometric constraints encountered in practice. Instead, even non-convex or disconnected feasible sets typically leave large regions entirely on one side of the boundary, as illustrated for the obstacle constraints in the SDD dataset in figure \ref{fig:sparsity}.

\begin{figure}[!t]
    {
    
    \setlength{\fboxsep}{0pt}
    \begin{center}
    \begin{minipage}{0.4\linewidth}
    \centering
    \includegraphics[
        width=\linewidth,
    ]{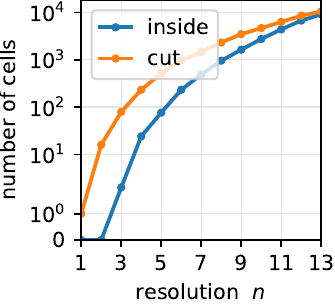}
    \end{minipage}
    \hspace{10pt}
    \begin{minipage}{0.4\linewidth}
    \centering
    \includegraphics[
        width=\linewidth,
    ]{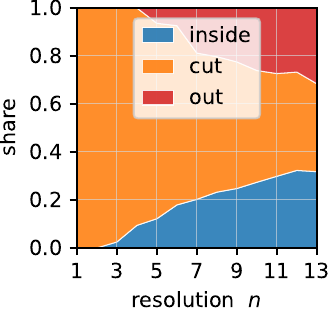}
    \end{minipage}
    \end{center}
    }
    \caption{\textbf{Exploiting sparsity limits expensive constrained integration to where it is needed.} As resolution increases, cells cut by the constraint boundary become increasingly sparse relative to fully feasible cells. We show their absolute numbers (left) and relative shares (right) for Full-Stokes (\ref{sec:full_stokes}). See \cref{app:s2_sparsity_plot} for details.}
    \label{fig:full_stokes_sparsity}
\end{figure}

This can lead to real speedups, especially in higher dimensions. We show this for the 4-dimensional Full-Stokes problem in figure \ref{fig:full_stokes_sparsity}. Here, one can see that when increasing the resolution per axis, the amount of cells fully inside the feasible rapidly approaches the number of cut cells. Exploiting this leads to $52\%$ the compile time and approximately half the memory pressure during evaluation compared to vanilla \gasp.

\subsection{S3: Adaptive Resolution}
\label{sec:gasp_s3}
\begin{figure}[h]
\vspace{-12pt}
    {
    \setlength{\fboxsep}{0pt}
    \begin{minipage}{0.35\linewidth}
    \centering
    \includegraphics[
        width=\linewidth,
    ]{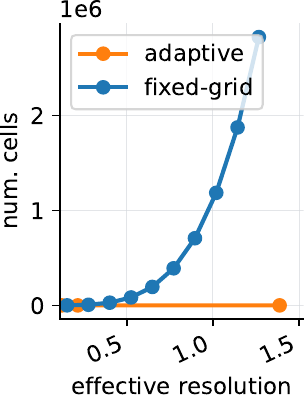}
    \end{minipage}
    }
    \hfill
    \begin{minipage}{0.6\linewidth}
    \caption{\textbf{Through adaptive resolution, \pcPal places a higher resolution where it matters and scales to high dimensions.} We show the relationship between the interpolation quality, as measured by the average ratio between the predicted gaussian scale and the cell-width to the number of pieces. Details in \cref{app:s3_plot_settings}.}
    \end{minipage}
    \label{fig:full_stokes_ball_figure}
\end{figure}
\begin{figure}[!t]
    {
    \setlength{\fboxsep}{0pt}
    \begin{minipage}{0.3\linewidth}
    \centering
    Gaussian\\[1pt]
    \fbox{%
    \includegraphics[
        width=\linewidth,
        clip
    ]%
{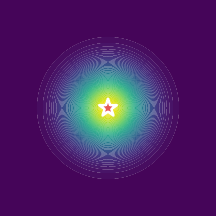}%
    }
    \end{minipage}
    \hfill
    \begin{minipage}{0.3\linewidth}
    \centering
    Fixed Grid\\[1pt]
    \fbox{%
    \includegraphics[
        width=\linewidth,
        clip
    ]{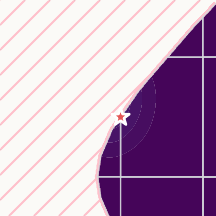}%
    }
    \end{minipage}
    \hfill
    \begin{minipage}{0.3\linewidth}
    \centering
    Hot Cells\\[1pt]
    \fbox{%
    \includegraphics[
        width=\linewidth,
        clip
    ]{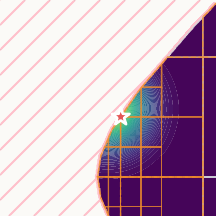}%
    }
    \end{minipage}
    }
    \hfill
    \caption{\textbf{By refining hot cells we increase resolution where it matters}. We statically analyze the which cells are commonly used and recursively refine them with a denser grid. We show unconstrained gaussian (left), coarse-cell (middle) and hot-cell (right) prediction for a Full-Stokes prediction (\ref{sec:full_stokes}).}
    \label{fig:hot_cell}
\end{figure}

While \cref{sec:gasp_s1,sec:gasp_s2} focused on scaling \pcPal and \gasp to higher dimensions, high dimensionality makes fine-grained interpolation prohibitively expensive. Accurate interpolation of sharp predictive densities requires small grid cells, yet the number of these cells grows exponentially with dimensionality. For example, on the Full-Stokes dataset, the average predicted gaussian scale $\sigma$ is $0.05$ but the domain is $[-1,1]^3{\times}[0,1]$. So for a resolution $\sigma/\texttt{width}=1.5$ we would need around $50$ pieces per axis which would result in 5.764.801 pieces. Computing the necessary moments beforehand times out after $3$ hours and computing the required coefficients would consume at least 55.3 GB of GPU-Memory for the resulting sparse cut-cell moments alone, which is infeasible. Crucially, we do not need this resolution everywhere. Predictions are usually not distributed uniformly across the high-dimensional space, but concentrate in smaller regions of the space. By adapting the grid to place a higher resolutions where it matters, we can therefore maintain accurate interpolation without an exploding the number of cells. Starting from a coarse grid, we analyze the training predictions to identify cells that are both frequently used and insufficiently resolved, which we call \emph{hot cells}. As illustrated in Figure~\ref{fig:hot_cell}, we recursively subdivide these cells until their local resolution adequately captures the predictions they contain. The resulting grid concentrates fine cells around sharp, frequently occurring predictions while remaining coarse where additional resolution provides little benefit. This dramatically reduces the number of cells required: an expected resolution of $1.5$ requires only $556$ cells, a reduction of more than $99.99\%$ compared with a uniform grid at the same resolution. See \cref{app:s3_algo} for details.

As we have now demonstrated how to exploit structure to scale \pcPal to high dimensional problems while staying efficient, exact and with high local resolution, we have everything we need to apply \pcPal to real-world problems.

\section{Related Work}
Many probabilistic NeSy methods~\citep{nesysurvey} enforce constraints only in expectation via loss penalties, using propositional~\citep{semanticloss1,semanticloss2}, fuzzy~\citep{sbr}, physical~\citep{DBLP:conf/aaai/StewartE17}, or algebraic~\citep{dl2} formulations. This typically yields no feasible sampling procedure and does not guarantee constraint satisfaction. Unlike \pcPal, these approaches therefore do not provide valid constrained distributions.
Neuro-symbolic approaches to sequential problems, such as time-series prediction, have so far received limited attention. A closely related line of work is relational neuro-symbolic Markov models (NeSy-MMs) \citep{DBLP:conf/aaai/SmetVRM25}. In contrast to our setting, NeSy-MMs focus on hidden Markov models with constraints imposed on the latent state space.%
While being able to express both discrete and continuous constraints using DeepSeaProblog \citep{DBLP:conf/uai/SmetMMMKR23},
in contrast to \pcPal, closed form constrained distributions are not available and inference relies on approximate NeSy particle filtering.
We expand upon the related work in the appendix in \cref{app:related_work}.

\section{\pcPal in action}
\label{sec:exps}

We now empirically evaluate \pcPal on two real-world applications, progressively testing its predictive benefits and scalability (C3). Specifically, we aim to answer two research questions: \textbf{Q1)} Can \pcPal enforce exact constraint satisfaction while retaining dependencies across the constraint factorization? \textbf{Q2)} Can \pcPal scale exact constrained log-likelihoods to high-dimensional problems while improving upon unconstrained predictions? We first evaluate \pcPal in an 8-dimensional trajectory-prediction setting, before scaling to a substantially higher-dimensional optical engineering application. Experimental details are provided in the appendix in \cref{app:experiments}.

\subsection{Trajectory Prediction}
\label{sec:exp_sdd}

In our first experiment, we investigate \textbf{Q1} in trajectory prediction on the Stanford Drone Dataset (SDD; \citealp{DBLP:conf/eccv/RobicquetSAS16}), a setting where decoupling the constraint and distribution factorizations is essential. While the constraints naturally decompose across time steps, trajectories themselves are strongly correlated across time. Imposing the same factorization on the predictive distribution would therefore fundamentally misrepresent the distribution.

To measure the ability of \pcPal to model tight correlations, we focus on the $4$-step prediction setting. Here, we observe five consecutive past positions of an agent and jointly predict its next four positions, yielding an 8-dimensional predictive distribution. The prediction steps are spaced further apart ($64$ frames vs $10$ frames) than the observations, requiring the model to capture substantial uncertainty and dependencies across the predicted trajectory. We impose the same semantic constraints as \citet{DBLP:conf/uai/KurscheidtMSPV25}, which restrict each predicted position based on the scene geometry and use all the $47$ scenes with recorded constraints. Appendix \ref{app:exp_sdd} details our setting.

\begin{figure*}[!t]
\centering
\setlength{\tabcolsep}{2pt}
\begin{minipage}{0.6\textwidth}
{
\setlength{\fboxsep}{0pt}
\begin{tabular}{cccc}
Step 1 & Step 2 & Step 3 & Step 4 \\
\fbox{%
\includegraphics[
width=0.22\linewidth,
trim={0 11 0 12},
clip
]{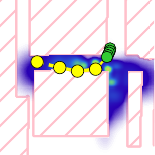}%
} &
\fbox{%
\includegraphics[
width=0.22\linewidth,
trim={0 11 0 12},
clip
]{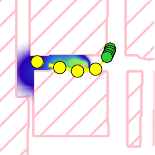}%
} &
\fbox{%
\includegraphics[
width=0.22\linewidth,
trim={0 11 0 12},
clip
]{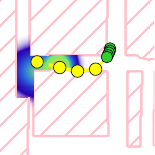}%
} &
\fbox{%
\includegraphics[
width=0.22\linewidth,
trim={0 11 0 12},
clip
]{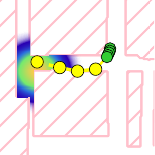}%
} \\
\fbox{%
\includegraphics[width=0.22\linewidth,
trim={0 11 0 12},
clip
]{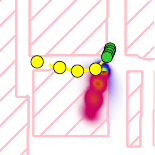}%
 } &
\fbox{%
\includegraphics[width=0.22\linewidth,
trim={0 11 0 12},
clip
]{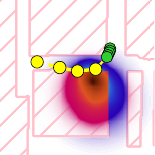}%
 } &
\fbox{%
\includegraphics[width=0.22\linewidth,
trim={0 11 0 12},
clip
]{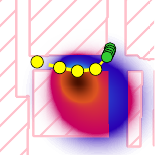}%
 } &
\fbox{%
\includegraphics[width=0.22\linewidth,
trim={0 11 0 12},
clip
]{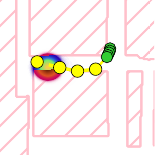}%
} \\
\end{tabular}
}
\end{minipage}
\begin{minipage}{0.38\textwidth}
\caption{\textbf{\pcPal captures the tight correlations between consecutive steps, while natively wrapping probability mass around the infeasible obstacles.} We show step-wise %
conditional predictions for a 4-step rollout, where each step is conditioned on all previous steps and the marginal distribution over all future steps is shown. 
Top row is NN+\pcPal and the bottom row is the NN+GMM(RNN).}
\label{fig:sdd_trajects_condition_expect}
\end{minipage}
\end{figure*}

For \pcPal, we represent the prediction using $450$ trajectory components, where the position at each time step is modeled as a linear combination of $30$ Gaussian position distributions. To obtain the polynomial densities required by \pcPal, we interpolate the univariate Gaussian inputs with squared splines, denoting the resulting family by $\tilde{\mathcal{N}}$. Concretely, the circuit is:
{%
\begin{equation*}
\begin{aligned}
    Y_{y}^{t} &= \pcProd(Y_{1, y}^{t},Y_{2, y}^{t}) & \text{30 position distributions}\\
    \vL^{t} &= \pcSum(\vY^{t}) & \text{projection to $450$ latents}\\
    \vL &= \pcProd(\vL^{t=1}, \dots, \vL^{t=4}) & \text{product over time}\\
    q &=  \pcSum(\vL) & \text{mixture over trajectories}\\
\end{aligned}
\end{equation*}
}%
with $Y_{i,y}^{t} \sim \tilde{\mathcal{N}}(\mu_{i,y}^{t}, \sigma_{i,y}^{t})$ denoting the $y$-th input distribution for spatial axis $i$ at time step $t$. The parameters of the circuit are not directly predicted, instead they are generated by an RNN \citep{DBLP:journals/corr/ChoMGBSB14} over the time-axis from a latent predicted by a ResNet \citep{DBLP:journals/corr/HeZRS15} with a final MLP \citep{Goodfellow-et-al-2016}, details are in \cref{sec:exp_sdd_pcpal}.

We compare \pcPal against three baselines, two of which replace the \pcPal distribution with a GMM \citep{370fbeadb5584ba9ab2938431fc4f140}. To closely match the capacity of \pcPal, both GMMs use $450$ mixture components: one directly predicts a full-covariance GMM, while the other uses the same recurrent parameterization as \pcPal with a banded covariance structure. All models receive the same inputs—the observed trajectory positions, the rasterized constraints, and the scene image. This setting highlights a key advantage of \pcPal: rather than requiring the predictive model to learn how to respect the feasible region from data, \pcPal incorporates its geometry directly into the predictive distribution. The model can therefore concentrate on learning the distribution over feasible trajectories, while constraint satisfaction is guaranteed by construction. The other baseline is the waypoint-conditioned Y-Net \citep{DBLP:journals/corr/abs-2012-01526}, a state of the art long-term trajectory prediction baseline. Results in table \ref{tab:sdd_trajects_comparison} show a significant improvement of \pcPal over both GMM and Y-Net in the log-likelihood as well as fidelity of the sampled trajectories. Here, ADE@$n$ denotes the minimum RSME over $n$ trajectories samples and FDE@$n$ denotes the minimum RSME over the $n$ sampled final position to the ground truth. Figure~\ref{fig:sdd_trajects_condition_expect} illustrates this advantage. We progressively reveal the given future trajectory and visualize the conditional distribution over the remaining positions,
$
p(\vY^{t\geq i}{\mid}\vY^{t<i}{=}\hat{\vy}^{t<i},\vx).
$
This shows how the predicted future changes as more of the trajectory becomes known.

While the GMM(RNN) captures strong correlations across time, the model is not aware that a correlated trajectory must also navigate the scene: its probability mass can connect likely positions by going through obstacles or off-road. It therefore assigns substantial probability to trajectories that are coherent across time but cannot actually occur. Only as more positions are revealed is the prediction guided toward the path taken by the agent. \pcPal, in contrast, wraps its probability mass around obstacles from the outset. It can represent multiple possible ways around an obstacle and, as more of the trajectory is revealed, progressively concentrate on the path actually taken.

Beyond raw predictive performance, \pcPal offers capabilities unavailable to standard probabilistic models: the same constraint language used to restrict the predictive distribution can also be used to manipulate and reason about it. This is illustrated in figure \ref{fig:sdd_semantic_queries}, where we highlight two applications: first, we can manipulate the predicted probability distribution, so for a prediction $P(\vY|\vx)$ we can restrict it to all paths that land inside some target polytope $\raisebox{-0.2em}{\includegraphics[height=1.1em]{figures/visualizations/emojii/checkered_flag-2.png}}$, i.e., compute $P(\vY|Y_4{\in}\raisebox{-0.2em}{\includegraphics[height=1.1em]{figures/visualizations/emojii/checkered_flag-2.png}}, \vx)$. This can be used to simulate scenarios, e.g. what if the agent turns right? %
More importantly, we can formulate safety questions directly over the entire predicted trajectory and answer them exactly. Suppose we are planning a trajectory $\vY_{\texttt{plan}}$ while another agent, whose future trajectory is described by $P(\vY\mid\vx)$, moves through the same space. We define a safety region $\raisebox{-0.2em}{\includegraphics[height=1.1em]{figures/visualizations/emojii/boom.png}}$ around our planned path and can directly ask: \emph{what is the probability that the other agent enters this region at any point in the future?} Since this event can be described by the same algebraic constraints, we can compute its probability $
P\left(\vY{\in}{\raisebox{-0.2em}{\includegraphics[height=1.1em]{figures/visualizations/emojii/boom.png}}}\mid\vx\right)
$ under the \pcPal prediction exactly. In the scenario shown in Figure~\ref{fig:sdd_semantic_queries}, this reveals a $43\%$ risk for the planned trajectory.

\begin{table}[ht]
    \caption{\textbf{\pcPal combines guaranteed constraint satisfaction and exact log-likelihoods with high-quality trajectory predictions.} We compare \pcPal against two GMM baselines and Y-Net, reporting log-likelihood (LL) and best-of-$n$ average/final displacement errors.}
    \label{tab:sdd_trajects_comparison}
    \begin{center}
    {
    \small
    \setlength{\tabcolsep}{3pt}
    \begin{tabular}{lccccc}
        \hline
        model & LL & ADE@1 & ADE@20 & FDE@1 & FDE@20\\
        \hline
Y-Net &	— & 27.31 & 13.91 & 36.88 & 16.47 \\
GMM & -23.23 & 40.58 & 12.65 & 67.88 & 19.13\\
GMM{\scriptsize(RNN)} & -21.88 & \textbf{16.87} & 7.97 & 29.37 & 8.95\\
\pcPal & \textbf{-21.55} & \textbf{16.87} & \textbf{7.44} & \textbf{26.71} & \textbf{7.80}\\
        \hline
    \end{tabular}
    }
    \end{center}
\end{table}

Together, these results demonstrate that exact constraint satisfaction need not come at the cost of distributional expressivity: \pcPal retains the strong cross-time dependencies required for realistic trajectory prediction while guaranteeing exact constraint satisfaction.

\subsection{Full Stokes}
\label{sec:full_stokes}

In order to demonstrate the scaling of \pcPal to high-dimensional problem, we focus on an application from optical engineering \citep{redding2013compact}. Here, the task is the following: the goal is to reconstruct polarization the polarization and intensity measurements in $81$ different wavelength-buckets from a deterministically mixed signal. These properties are described by the four Stokes parameters $(S_0,S_1,S_2,S_3)$. Here, $S_0$ measures the total light intensity, while $S_1$ and $S_2$ measure preferences for different directions of linear polarization and $S_3$ captures circular polarization. Together, $\Vec{\vs}{=}(S_1,S_2,S_3)$ therefore describe how strongly and in which way the light is polarized. Physically, this polarized component cannot be stronger than the total light intensity, giving the constraint
$
S_1^2{+}S_2^2{+}S_3^2\ {\leq}\ S_0^2
$
per wavelength bucket \citep{fullstokesineq}. Since these constraints are not in \smtlra, we over approximate these constraints as the over-approximation results in \pcPal computing a lower bound of the true constrained likelihood (compare \cref{app:full_stokes_cone_constraints}).

\begin{figure}[!t]
    {
    \setlength{\fboxsep}{0pt}
    \begin{minipage}{0.3\linewidth}
    \centering
    $\Vec{\vs}{\sim}\mathcal{N}(\mu,\sigma)$\\unconstrained\\[1pt]
    \includegraphics[
        width=\linewidth,
        trim={0 20 0 20},
        clip
    ]{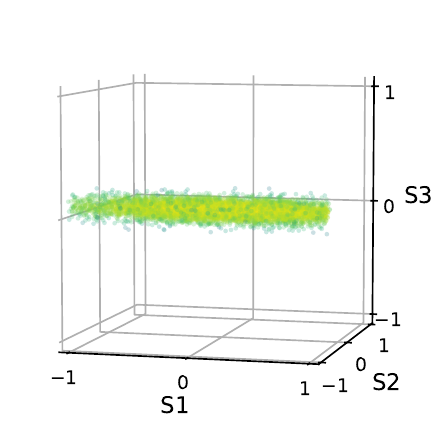}
    \end{minipage}
    \hfill
    \begin{minipage}{0.3\linewidth}
    \centering
    $\Vec{\vs}{\sim} \tilde{\mathcal{N}}(\mu,\sigma)$\\under $\lVert \Vec{\vs} \rVert_2^2 \leq S_0^2$\\[1pt]
    \includegraphics[
        width=\linewidth,
        trim={0 20 0 20},
        clip
    ]{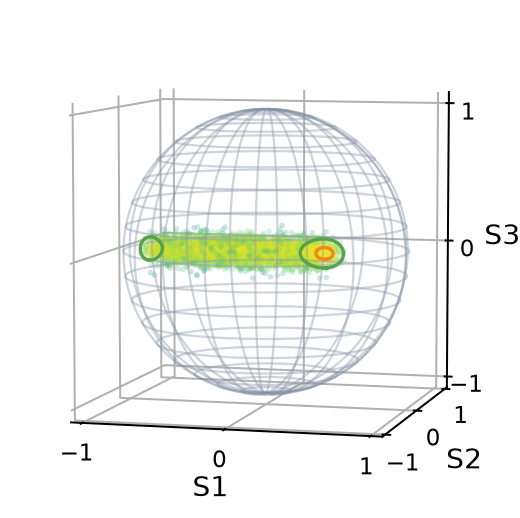}
    \end{minipage}
    \hfill
    \begin{minipage}{0.3\linewidth}
    \centering
    $\Vec{\vs}{\sim} \tilde{\mathcal{N}}(\mu,\sigma)$\\under $\lVert \Vec{\vs} \rVert_2^2 = 1$\\[1pt]
    \includegraphics[
        width=\linewidth,
        trim={0 20 0 20},
        clip
    ]{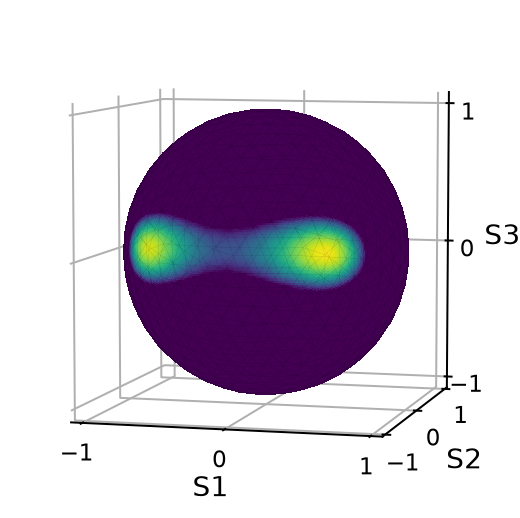}
    \end{minipage}
    }
    \caption{\textbf{\pcPal enables flexible composition with constraints}. %
    On the left, we show the unconstrained gaussian distribution, in the middle the constraints to the unit-ball, and on the right the specific constraints for the polar-dataset, where the dataset lies on the surface.}
    \label{fig:full_stokes_ball_figure}
\end{figure}

\begin{table}[h]
    \caption{\textbf{\pcPal improves predictions by composition with constraints.} Without changing the underlying predictor, \pcPal improves both in log-likelihood and prediction accuracy.}
    \label{tab:full_stokes_comparison}
    \begin{center}
    {
    \setlength{\tabcolsep}{4pt}
    \begin{tabular}{lccc}
        \toprule
        Model & LL & rel. MAE & rel. $\Vec{\vs}$-MAE\\
        \midrule
FullStokes & 478.86 & 421.37 & 369.49\\
FullStokes + DLR & - & 408.60 & 312.64 \\
FullStokes + \pcPal & \textbf{590.89} & \textbf{347.51} & \textbf{267.07}\\
        \bottomrule
    \end{tabular}
    }
    \end{center}
\end{table}

In order to apply \pcPal to the problem, we first train the point-predictor by reproducing the model from \citet{redding2013compact}. We then freeze this model and extend it with an additional output predicting the scale $\sigma$, turning the original point prediction into a Gaussian predictive distribution. Using our $\tilde{\mathcal{N}}$ construction, \pcPal then incorporates an algebraic over-approximation of the physical constraints into the predictive distribution. In Table~\ref{tab:full_stokes_comparison}, we compare the resulting \pcPal model against the unconstrained Gaussian prediction and an \smtlra-constrained projection based on the Disjunctive Linear Refinement (DLR) layer \citep{DBLP:conf/iclr/StoianG25}. As shown in Table~\ref{tab:full_stokes_comparison}, \pcPal improves both log-likelihood and MAP (via PaMAP \citep{DBLP:journals/corr/abs-2602-08681}, see \cref{app:full_stokes_map}) over the unconstrained Gaussian and the projection-based baseline. Notably, the Gaussian MAP corresponds to the original point predictor, showing that \pcPal improves the point prediction purely by incorporating the constraints. This positively answers \textbf{Q2} and demonstrates that \pcPal scales to high-dimensional real-world problems.

Finally, \pcPal is not restricted to full-dimensional constrained spaces. We demonstrate this flexibility on the single-wavelength Full-Stokes dataset for bucket $40$, where the intensity is fixed to $S_0=1$ and the light is almost fully polarized (abs. deviation $0.01$). This provides stronger problem-specific knowledge than the general Stokes constraints: fixing $S_0{=}1$ and assuming full polarization restricts $(S_1,S_2,S_3)$ to the surface of the unit ball. We approximate this surface by a triangulation (Figure~\ref{fig:full_stokes_ball_figure}, see \cref{app:full_stokes_surface_constraints} for more details) and use it to constrain the \pcPal prediction. Crucially, we impose this additional structure only through \pcPal: the underlying model continues to predict the full four-dimensional Stokes vector. Composing this prediction with the more specific constraints restricts its support from the four-dimensional ambient space to a two-dimensional triangulated surface. Despite this change in dimensionality, \pcPal retains an exact density on the resulting representation. This more informative constraint improves test performance over both the unconstrained baseline and \pcPal with the general Stokes constraints (Table~\ref{tab:full_stokes_ball_comparison}).

This demonstrates the compositional nature of \pcPal particularly clearly: changing our knowledge about the problem does not require changing the predictive model. Instead, we can compose the same four-dimensional prediction with a different constraint set—even one that changes the dimensionality of its support. Here, additional domain knowledge turns the same prediction into a distribution on a two-dimensional surface without requiring a problem-specific reparameterization of the model.

\begin{table}[t!]
    \caption{\textbf{\pcPal composition adapts the same predictor to more specific domain knowledge.} Replacing the general 4D constraints with the task-specific 2D surface substantially improves point-prediction accuracy.}
    \label{tab:full_stokes_ball_comparison}
    \begin{center}
    {
    \setlength{\tabcolsep}{4pt}
    \begin{tabular}{lccc}
        \hline
        Model & LL & rel. MAE & rel. $\Vec{\vs}$-MAE\\
        \hline
        \multicolumn{4}{l}{\textit{4D distribution}}\\
Full Stokes & -309.55 & 28.07 & \textbf{23.44}\\
Full Stokes + DLR & - & 28.02 & 23.93\\
FullStokes + \pcPal &  \textbf{-241.09} &\textbf{ 27.79} & {23.51}\\
\hline
\multicolumn{4}{l}{\textit{2D surface distribution}}\\
FullStokes + \pcPal &  \textbf{-45.51} & \textbf{17.06} & \textbf{22.75}\\
        \hline
    \end{tabular}
    }
    \end{center}
\end{table}

\section{Conclusion}

In this work, we introduced \pcPal, a compositional and scalable approach to probabilistic machine learning under algebraic constraints that exploits structure found in many applications. Across two real-world applications, trajectory prediction and optical engineering, \pcPal improves log-likelihood as well as sampling- and MAP-based predictions, demonstrating that expressive uncertainty modeling and accurate predictions can be combined with guaranteed constraint satisfaction. Beyond predictive performance, we demonstrate the flexibility of \pcPal: constraints can be used to condition predictions on semantic events, compute probabilities of structured events, and refine the distributions support as more specific domain knowledge becomes available. %
Looking forward, combining our compile-time adaptive refinement with runtime refinement could extend \pcPal to less concentrated predictive distributions, opening applications such as fluid simulation \citep{ohana2024well} where predictions span larger regions of the space.

\subsection*{Acknowledgements}
We thank Nicola Branchini and Adri{\'a}n Javaloy for valuable feedback on the draft. AV was supported by the “UNREAL: Unified Reasoning Layer for Trustworthy ML” project (EP/Y023838/1) selected by the ERC and funded by UKRI EPSRC, and acknowledges funds from Huawei TTE-DE Lab Munich.

\subsection*{Contribution}
LK conceived the initial idea of the paper. LK is responsible for all theoretical contributions, illustrations, algorithms, implementation and experiments. LK wrote the paper with help from AV. AV supervised all phases of the project.

\subsection*{AI use statement}
In this work, we used generative AI tools for assisting in coding and to refine the writing. LLM-generated code was verified and tested for correctness.  
We take responsibility for the final content of this work, including text, claims, or artifacts produced with the aid of generative AI.

\bibliography{full_paper_bib}

\section*{Checklist}

\begin{enumerate}

  \item For all models and algorithms presented, check if you include:
  \begin{enumerate}
    \item A clear description of the mathematical setting, assumptions, algorithm, and/or model. [\textbf{Yes}/No/Not Applicable]
    \item An analysis of the properties and complexity (time, space, sample size) of any algorithm. [\textbf{Yes}/No/Not Applicable]
    \item (Optional) Anonymized source code, with specification of all dependencies, including external libraries. [Yes/\textbf{No}/Not Applicable]\\ \textbf{Source Code will be released upon acceptance.}
  \end{enumerate}

  \item For any theoretical claim, check if you include:
  \begin{enumerate}
    \item Statements of the full set of assumptions of all theoretical results. [\textbf{Yes}/No/Not Applicable]
    \item Complete proofs of all theoretical results. [\textbf{Yes}/No/Not Applicable]
    \item Clear explanations of any assumptions. [\textbf{Yes}/No/Not Applicable]
  \end{enumerate}

  \item For all figures and tables that present empirical results, check if you include:
  \begin{enumerate}
    \item The code, data, and instructions needed to reproduce the main experimental results (either in the supplemental material or as a URL). [Yes/\textbf{No}/Not Applicable]\\ \textbf{Source Code will be released upon acceptance.}
    \item All the training details (e.g., data splits, hyperparameters, how they were chosen). [\textbf{Yes}/No/Not Applicable]
    \item A clear definition of the specific measure or statistics and error bars (e.g., with respect to the random seed after running experiments multiple times). [\textbf{Yes}/No/Not Applicable]
    \item A description of the computing infrastructure used. (e.g., type of GPUs, internal cluster, or cloud provider). [\textbf{Yes}/No/Not Applicable]
  \end{enumerate}

  \item If you are using existing assets (e.g., code, data, models) or curating/releasing new assets, check if you include:
  \begin{enumerate}
    \item Citations of the creator If your work uses existing assets. [\textbf{Yes}/No/Not Applicable]
    \item The license information of the assets, if applicable. [\textbf{Yes}/No/Not Applicable]
    \item New assets either in the supplemental material or as a URL, if applicable. [\textbf{Yes}/No/Not Applicable]
    \item Information about consent from data providers/curators. [\textbf{Yes}/No/Not Applicable]
    \item Discussion of sensible content if applicable, e.g., personally identifiable information or offensive content. [\textbf{Yes}/No/Not Applicable]
  \end{enumerate}

  \item If you used crowdsourcing or conducted research with human subjects, check if you include:
  \begin{enumerate}
    \item The full text of instructions given to participants and screenshots. [Yes/No/\textbf{Not Applicable}]
    \item Descriptions of potential participant risks, with links to Institutional Review Board (IRB) approvals if applicable. [Yes/No/\textbf{Not Applicable}]
    \item The estimated hourly wage paid to participants and the total amount spent on participant compensation. [Yes/No/\textbf{Not Applicable}]
  \end{enumerate}

\end{enumerate}

\clearpage
\appendix
\thispagestyle{empty}

\onecolumn
\aistatstitle{Supplementary Material}

\section{Exploiting Structure}
\label{app:exploiting_structure}
\subsection{S1: Scaling up \gasp}
\label{app:s1}

\subsection{\gasp Benchmark}
\label{app:s1_gasp_benchmark}
This benchmarks generate a varying number of two-dimensional groups connected via constraints. Each group is connected by generating two non-overlapping, two-dimensional simplices. We generate benchmark data from $1$ to $10$ groups. We compare the time it takes to compute all the constrained moments for a decomposable polynomial of degree $6$ per dimension using \gasp for the enumeration of \pal (not exploiting decomposability) vs. \pcPal (which exploits the decomposability). We run the benchmark on a NVIDIA L40-GPU with a timeout of $24$ hours per group.

\subsubsection{Tractability WMI}
\label{app:tractability_wmi}

\begin{proposition}[Tractability of decomposable WMI]
\label{prop:decomposable_wmi}
Let
\[
\formula(\vY)=\bigwedge_{i=1}^{m}\formula_i(\Ygroup[i])
\]
with pairwise disjoint groups $\Ygroup[i]$ of bounded dimensionality,
$\max_i |\Ygroup[i]|\leq d$ for constant $d$.
Assume that:
\begin{enumerate}
\item $q(\vY)$ has bounded degree and decomposes over the same groups through sums and products;
\item AllSAT enumeration for each $\formula_i$ produces polynomially many convex polytopes in polynomial time; and
\item each resulting polytope has a polynomial-size linear description.
\end{enumerate}
Then the weighted model integral
\[
\int q(\vy)\Ind{\vy\models\formula}\,d\vy
\]
can be computed in polynomial time.
\end{proposition}

\begin{proof}
Since $q$ decomposes through sums and products over disjoint variable groups, linearity of integration and Fubini's theorem allow the integral to be propagated through the same decomposition. It therefore suffices to compute integrals of the form
\[
\int q_i(\vy_i)\Ind{\vy_i\models\formula_i}\,d\vy_i
\]
for individual groups $\Ygroup[i]$.

By assumption, AllSAT enumeration decomposes the feasible set of each $\formula_i$ into polynomially many convex polytopes. Each polytope is described by polynomially many linear inequalities and has dimension at most the constant $d$. In fixed dimension, such a polytope admits a triangulation into polynomially many simplices \citep{ziegler1993lectures}. Hence, each group-wise WMI reduces to polynomially many integrals of bounded-degree polynomials over fixed-dimensional simplices.

Integration of a bounded-degree polynomial over a simplex in fixed dimension is computable in polynomial time \citep{baldoni2011integrate}. Each group-wise integral is therefore polynomial-time computable. Since there are at most polynomially many groups and the decomposition of $q$ combines their integrals using only polynomially many additions and multiplications, the complete WMI is polynomial-time computable.
\end{proof}

\subsubsection{\pcPal}
\label{app:pc_pal}

We will now explain in more detail how \pcPal allows us to exploit the factorization structure \cref{eq:decomp}. Let us consider again how the circuit structure splits the integral in e.g. for a simple mixture model:
\begin{align}
    l_i' &:=p_i'(y_1) && \text{($i$-th polynomial input density over $x_1$)} \\
    l_i'' &:=p_i''(y_2) && \text{($i$-th polynomial input density over $x_2$)}\\
    \vl &= \pcProd(\vl', \vl'') && \text{(element-wise product)}\\
    \tilde{p}(y_1,y_2) &= \pcSum(\vl) && \text{(unnormalized density $\tilde{p}$)}
\end{align}
We assume coefficients in $\pcSum$ in order to guarantee the non-negativity of $\tilde{p}$.

We will now assume $y_1$ to have domain $A$ and $y_2$ to have domain $B$. If we want to compute the normalizing constant, we can ``push'' the integral back to the input functions and evaluate the circuit on the input-functions integral instead of its point-wise density. In order to stress that the value of the latents is a function of $x$, we write $\vl|_{\vy=\vy'}$ to denote the value of latent $\vl$ at point $\vy'$, in addition we interpret the integral over a vector of functions as the element-wise integral:
\begin{align}
    \int_{A\times B} \tilde{p}(\vy') d\vy'&= \int_{A\times B} \pcSum(\vl|_{\vy=\vy'})  d\vy' \\
    &= \pcSum( \int_{A\times B}\vl|_{\vy=\vy'}d\vy') && \text{(linearity of $\pcSum$)}\\
    &= \pcSum( \int_{A\times B}\pcProd(\vl'|_{y_1=y_1'}, \vl''|_{y_2=y_2'})d\vy')\\
    &= \pcSum(\pcProd(\int_{A}\vl'|_{y_1=y_1'}dy_1, \int_{B}\vl''|_{y_2=y_2'}dy_2)) && \text{(decomposition)} \label{eq:circuit_decomp}\\
    &= \pcSum(\pcProd(\int_{A}[p_i'(y_1)]_{i=1}^n dy_1, \int_{B}[p_i''(y_2)]_{i=1}^n dy_2)) && \text{(integral over input distributions)}
\end{align}
In the unconstrained case, we can always exploit the circuit structure for our considered circuit-topologies to compute the normalizing constant \citep{vergari2019tractable}. A crucial requirement is that whenever we want to decompose the integral over $\pcProd(\vl, \vl')$ with variables $x_{left}:=\mathrm{sc}(\vl)$ and $x_{right}:=\mathrm{sc}(\vl')$ into integrals over $\vl$ and $\vl'$, the domain $D$ over which the integral is transformed similarly decomposes. So the domain $D$ is spanned by the cartesian product $A\times B=D$, with $A$ being a set over $y_{\mathrm{left}}$ and $B$ over $y_{\mathrm{right}}$.

We recall our factorization assumption:
\begin{equation}
\formula(\vY)
= \bigwedge\nolimits_i \formula_i(\Ygroup[i]),
\end{equation}

A domain always arises from a cartesian product if the constraint decomposes:
\begin{align}
    D &= \{\vy \mid \vy \models \formula \} \\
    &= \{\vy \mid (\vy_{\mathrm{left}} \models \formula_{\mathrm{left}}) \land (\vy_{\mathrm{right}} \models \formula_{\mathrm{right}})\} \\
    &= \{\vy_{\mathrm{left}} \mid \vy_{\mathrm{left}} \models \formula_{\mathrm{left}}\} \times \{\vy_{\mathrm{right}} \mid (\vy_{\mathrm{right}} \models \formula_{\mathrm{right}})\}
\end{align}

As a result, we can exploit the cirucit structure as long as the splits of the circuits are aligned with the decomposition of the formula. In this way, we can decompose the integral without factorizing the distributions, until we hit an ``atomic'' group $\Ygroup[i]$ that forbids us from decomposing the integral further.

If we arrive at an sub-circuit over an ``atomic'' group $\Ygroup[i]$, we can not exploit the circuit structure further to calculate its integral. We need to calculate the integral of the sub-circuit over $\formula_i(\Ygroup[i])$. But since we assume polynomial input-distributions and the circuit arises from repeatedly applying multi-linear functions, the resulting function modeled by the circuit is just a polyomial (or vector of polynomials). If we consider this circuit as one polynomial in a factorized form, we can normalize its representation to turn it into polynomial standard form and apply the \pal-approach to calculate its integral over $\formula_i(\Ygroup[i])$. This is a perfectly vectorizable step, both obtaining the standard form and the \pal-integral, and therefore we can compute the integral for the whole vector of polynomials in parallel. As the dimensionality of the $\Ygroup[i]$ is assumed bounded, this is both efficient and exact. After obtaining the integrals needed we can evaluate the rest of the circuit as usual to calculate the overall normalizing constant.

As an example, we can consider the circuit for the trajectory-prediction application (\cref{sec:exp_sdd}). The task is to predict a joint $4$-step trajectory-distribution $(\vY^{1}, \dots, \vY^{4})$ under obstacle constraints. We have the following circuit-topology:
\begin{align}
    Y_{i, y}^{t} &\sim \tilde{\mathcal{N}}(\mu_{i, y}^{t}, \sigma_{i, y}^{t}) & \text{spline interpolation of gaussian density}\\
    Y_{y}^{t} &= \pcProd(Y_{1, y}^{t},Y_{2, y}^{t}) & \text{30 $2$d-position distributions}\label{eq:sdd_circuit_2d_prod}\\
    \vL^{t} &= \pcSum(\vY^{t}) & \text{projection (mixture) to $450$ latents}\label{eq:sdd_circuit_to_latents}\\
    \vL &= \pcProd(\vL^{t=1}, \dots, \vL^{t=4}) & \text{element-wise product over time}\\
    q &=  \pcSum(\vL) & \text{mixture over trajectories}\label{eq:sdd_circuit_final_sum}
\end{align}
The constraints decompose as follows:
\begin{equation}
\formula_{\mathrm{SDD}}(\vY^{1}, \dots, \vY^{4})
= \bigwedge\nolimits_{i=1}^4 \formula_{\mathrm{Obstacles}}(\vY^{t=i})
\label{eq:app_sdd_constraint}
\end{equation}

In order the compute the normalizing constant we can again decompose the integral $\int_{\vy' \models \formula_{\mathrm{SDD}}} p_{SDD}(\vy') d\vy'$ according to the circuit structure until we hit the constraints ``binding'' variables $Y_{1, y}^{t}$ and $Y_{2, y}^{t}$ together. In our circuit, we can follow operations \ref{eq:sdd_circuit_final_sum} to \ref{eq:sdd_circuit_to_latents} to propagate the integral and only block at \ref{eq:sdd_circuit_2d_prod}.

The resulting algorithm to compute the normalizing constant for the trajectory-prediction circuit is therefore:
\begin{align}
    Y_{i, y}^{t} &\sim \tilde{\mathcal{N}}(\mu_{i, y}^{t}, \sigma_{i, y}^{t}) & \text{spline interpolation of gaussian density}\\
    \vY_{y}^{t} &= \pcProd(Y_{1, y}^{t},Y_{2, y}^{t}) & \text{30 polynomial $2$d-position distributions at time $t$}\\
    I_{y}^{t} &= \int p(\vY_{y}^{t}{=}\vy') \Ind{\vy' \models \formula_{\mathrm{Obstacles}}} d\vy' &\text{30 times \ref{eq:pal_wmi} at time $t$}\\
    \vI^{t} &= \pcSum(I_{y}^{t}) & \text{integral of $450$ latents per time-step}\label{eq:sdd_circuit_to_latents_integral}\\
    \vI &= \pcProd(\vI^{t=1}, \dots, \vI^{t=4}) & \text{integral of the $450$ trajectory-distributions, aligned with \ref{eq:app_sdd_constraint}}\\
    I &=  \pcSum(\vI) & \text{integral of the final Mixture}\label{eq:sdd_circuit_final_sum_integral}
\end{align}

We therefore managed to exploit the decomposition of the constraints to compute the normalizing constant without factorizing the predicted distribution accordingly.

\subsection{S2: Exploiting Sparseness}
\label{app:s2}

\subsubsection{Compiling And Evaluating With Sparse Cells}
\label{app:s2_sparse_algo}

We will now provide the algorithm to compile with a sparse support in \cref{alg:sparse_compiler}. The idea behind the algorithm is that, for a piecewise function $f$ with $h_i$ pieces per dimension, we return the integrated moments in a sparse datastructure $M_i=(M_i, \texttt{index}_i)$, where $M_i$ are the integrated moments and the $\texttt{index}_i$ the cell-index, and a tensor $I \in \mathbb{R}^{h_1 \times \dots h_n}$ with $I[\texttt{index}] = 1 \Leftrightarrow $ Cell \texttt{index} is fully feasible.

\begin{algorithm}[h]
\caption{Sparse Compilation (per group)}
\label{alg:sparse_compiler}
\begin{algorithmic}[1]
\Require \smtlra \formula, \text{piecewise, separable polynomial} $f_{\theta}:\mathbb{R}^n \rightarrow \mathbb{R}$
\Ensure \text{integrated basis functions} $M_i=(M_i, \texttt{index}_i)$, interior cell index $I$
\State $s \gets \Call{ShapeGrid}{f}$
\State $I \gets \Call{ZeroTensor}{s}$
\State $M \gets []$
\For{$(\texttt{shape}, \texttt{idx}) \in \Call{EnumerateCells}{f}$}
    \If{\Call{FullyFeasible}{\texttt{shape}, \formula}} \Comment{via SMT Solver}
        \State $I[\texttt{idx}] = 1$
    \Else
        \State $M_i \gets \Call{WMI}{f, (\texttt{shape} \wedge \formula)}$
        \State $\Call{Append}{M, (M_i, \texttt{idx})}$
    \EndIf
\EndFor
\State \Return $M, I$
\end{algorithmic}
\end{algorithm}

The algorithm can be sped up by detecting whenever $\formula$ is a polytope, as the cell is then fully feasible whenever the boundary points are fully feasible, and the WMI-integral is then just a single  $\Call{GaspIntegral}$ over the polytope $(\texttt{shape} \wedge \formula)$.

\begin{algorithm}[h]
\caption{Sparse Evaluation (per group)}
\label{alg:sparse_evaluator}
\begin{algorithmic}[1]
\Require $f_{\theta}:\mathbb{R}^n \rightarrow \mathbb{R}$, parameters $\Omega$, $M$, $I$
\Ensure Integral $F$ for $f_{\theta=\Omega}$ over $\formula$
\State $\hat{I}_i \gets \Call{UnivariateIntegrals}{i, f, \Omega}$ \Comment{$\hat{I}_i$ is a vector of length $h_i$ containing the integrals over the intervals}
\State $\hat{M} \gets 0$
\For{$(M_i, \texttt{idx}) \in M$}
    \State $m \gets \Call{DenseIntegral}{M_i, \texttt{idx}, f, \Omega}$
    \State $\hat{M} \gets \hat{M} + m$
\EndFor
\State $\hat{I} \gets \Call{TensorContraction}{I, hat{I}_i}$
\State \Return $\hat{I} + \hat{M}$
\end{algorithmic}
\end{algorithm}

While the sparse representation might seem inefficient to evaluate at first, the size of the moment tensor $M_i$ and the work in $\Call{DenseIntegral}$ dominate the memory transfer time, such that a slowdown is not observable. In comparison, the integral over the interior is fast, exact, without significant memory pressure and easy to implement on the GPU.

\subsubsection{Sparsity Overview Setting}
\label{app:s2_sparsity_plot}
We use the Full-Stokes constraints (see \cref{app:full_stokes_cone_constraints}) with the grid-resolution being parametrized univariate, so a resolution of $5$ means $5$ pieces per dimension. We run the benchmark on a NVIDIA L40-GPUs.

\subsection{S3: Adaptive Resolution}
\label{app:s3}

\subsubsection{Adaptive Resolution Overview Setting}
\label{app:s3_plot_settings}
We use the Full-Stokes constraints (see \cref{app:full_stokes_cone_constraints}) and compare an equi-spaced grid to an equi-spaced grid enhanced with the adaptive resolution fitted from the model using hot-cell algorithm with $K=n$ and $R=n$. We measure the effective resolution per cell as the geometric mean between the per-axis cell-width at the predicted mean divided by the predicted gaussian scale for the axis. This ratio is a good proxy for how well the spline can interpolate the gaussian density. We then compute this average over the test-set predictions. We run the benchmark on a NVIDIA L40-GPUs.

\subsubsection{Adaptive Resolution Algorithm}
\label{app:s3_algo}
The hot-cell algorithm works via two hyperparameters: $K$, so how many coarse cells we refine in the first level, and $R$, so how many hot-cells from the first-level we refine again. Since the hot-cells is data-dependent, we calculate it over the predictions over the validation-dataset.

\begin{algorithm}[h]
\caption{HotCellHelper}
\label{alg:hot_cell_algorithm_helper}
\begin{algorithmic}[1]
\Require coarse grid $g$, $n$, predicted distributions $\hat{p}$
\Ensure adaptive grid $a$, replaced $b$
\State $g_{mass} \gets \Call{ProbabilityMassPerCell}{g, \hat{p}}$
\State $g_{sorted} \gets \Call{Sort}{g, \texttt{via}=g_{mass}}$
\State $g_{refined} \gets \Call{Refine}{g_{sorted}[1{:}n]}$ \Comment{Subdivides every cell into $2^d$ children via split in per-axis middle}
\State \Return $g_{refined}$, $g_{sorted}[1{:}n]$
\end{algorithmic}
\end{algorithm}

\begin{algorithm}[h]
\caption{HotCells}
\label{alg:hot_cell_algorithm}
\begin{algorithmic}[1]
\Require coarse grid $G$, $K$, $R$, predicted distributions $\hat{p}$
\Ensure adaptive grid $A$ without coarse grid $G$
\State $A^1, B^1 \gets \Call{HotCellHelper}{G, K, \hat{p}}$
\State $A^2, B^2 \gets \Call{HotCellHelper}{A^1, R, \hat{p}}$
\State \Return $(A^2 \cup (A^1 / B^2))$
\end{algorithmic}
\end{algorithm}

After integrating over the cells, the resulting adaptive grid can be evaluated analogue to \cref{alg:sparse_evaluator}. We just have to track the level (coarse, refined, recursed into refined) to obtain the correct polynomial coefficients coefficients for the cell-index.

\section{Related Work}
\label{app:related_work}

Closely related to our work is work on constrained sampling. This includes Hit-and-Run\citep{smith1984efficient} based approaches, which recently got extended do flow-matching models\citep{DBLP:journals/corr/abs-2606-13400} and to incorporate higher-order information\citep{paul2026higher}. Another approach that gathered recent interested is mirror-based constrained handling in the context of diffusion models \citep{liu2023mirror} and landing-based methods\cite{jeon2026efficient}.

From a technique-perspective, our work is closely related to semantic probabilistic layers (SPL) \citep{ahmed2022spl}, which combine neural networks with probabilistic circuits but only for propositional logic. Similarly to SLP, DeepProbLog \citep{DBLP:conf/nips/ManhaeveDKDR18} combine probabilistic logic programming with neural networks, but also only for only Boolean logic and fully-factorized distributions.

\section{Experiments}
\label{app:experiments}

\subsection{Trajectory Prediction}
\label{app:exp_sdd}

We use the $47$ scenarios of the annotated SDD-dataset \citep{DBLP:conf/uai/KurscheidtMSPV25} for which we have trajectories long enough for 5 observed trajectory points with a $10$ frame difference and $4$ trajectory points to predict with a $64$ frame difference. We split a trajectory into multiple datapoints via a sliding window with a $145$ stride. We then split the dataset into $70\%$ train, $15\%$ validation and $15\%$ test ($10238$, $2192$ and $2193$). We are careful not to place datapoints derived from a single trajectory into multiple splits. Crucially, the $15\%$ test-set is further sub-divided ($1103$ vs $1090$) into one half with shared scenes, which we report in the main paper, and one half of trajectories from OOD-scenes ($2$-$6$) that do not appear in the other splits, which we use to verify that the constraint-satisfaction successfully transfers to new, unseen scenes in \cref{sec:exp_sdd_ood}. The scene image and rasterized constraints are provided at a $64\times64$ resolution, adjusted for the different aspect ratios, and concatenated as input. This gives the GMM baselines explicit access to the constraint information, providing a stronger and fairer comparison with \pcPal. Y-Net receives its inputs in the format prescribed by its original architecture; importantly, its architecture includes a pretrained semantic segmentation network, providing it with semantic information about the scene.

All architectures, including the Y-Net are trained using flipping and rotation as data-argumentation.

\subsubsection{Common Architecture for \pcPal and GMM}
\label{sec:exp_sdd_pcpal}

Both architectures start with the image-encoder, which is a ResNet \citep{DBLP:journals/corr/HeZRS15} with channel sizes $[32,64,128,256]$ leading to a $256$-sized latent. Then, we have the trajectory-encoder. Here, we have different sizes:
\begin{itemize}
    \item small: $[512, 512]$
    \item medium : $[1024, 1024]$
    \item large : $[2048, 2048]$
\end{itemize}
We then have a fusion MLP $[256, 256]$ that combines both inputs, and then projects it with a final linear layer to a $2048$-dim embedding. On this embedding, we run a 2-layer GRU \citep{DBLP:journals/corr/ChoMGBSB14} with hidden size $1536$ for the RNN-parametrized version or directly predict the coefficients for the non-RNN parametrized GMM.

We optimize using Adam.

\subsubsection{\pcPal}
\label{sec:exp_sdd_pcpal}
The circuit chosen is:
\begin{align*}
    Y_{y}^{t} &= \pcProd(Y_{1, y}^{t},Y_{2, y}^{t}) & \text{30 position distributions (selected by grid search)}\\
    \vL^{t} &= \pcSum(\vY^{t}) & \text{projection to $450$ latents}\\
    \vL &= \pcProd(\vL^{t=1}, \dots, \vL^{t=4}) & \text{product over time}\\
    q &=  \pcSum(\vL) & \text{mixture over trajectories}\\
\end{align*}

The input-distributions are gaussians, which we interpolate using squared splines (by interpolating the root of the gaussian density). We have $34$ equi-distant spline knots and the gaussian have a minimum-scale of $0.165$ (the trajectory domain is $[0,10]^2$). The gaussians are parametarized relative to the last observed position.

We also adjust the integral-computation. Numerical errors in the constrained moments are particularly problematic during training: while negligible for a fixed prediction, optimization can systematically exploit these errors by choosing coefficients that amplify them. In particular, numerical inaccuracies can introduce small negative directions into an integral that is theoretically non-negative, allowing the optimization objective to spuriously reduce the computed probability mass. To prevent this behavior, we evaluate the integral through an explicitly positive-semidefinite Gram representation:
\begin{align}
\texttt{unstabilized}(\alpha_1,\alpha_2,M)
    &= \alpha_1^\top M \alpha_2,\\
\texttt{gramStabilizedMoment}(\alpha_1,\alpha_2,M)
    &= \lVert F^T\alpha \rVert_2^2 \\[2pt]
G_{(a,b),(c,d)}
    &:= M_{a+c,b+d},
    && \text{Gram matrix (theoretically psd)},\\
FF^\top &= G_+,
    && \text{Decomposition $G$ with negative EV clipped},\\
    \alpha &= \alpha_1 \odot \alpha_2 \in \mathbb{R}^16 && \text{Combination of Coefficients}
\end{align}
where $\alpha_1$ and $\alpha_2$ are the coefficients for the polynomial over $y_1$ and $y_2$ and $M \in \mathbb{R^{7\times7}}$ is the constrained-moment Matrix. $\texttt{unstabilized}$ is the reference, unstabilized computation that directly arises from the constrained moment matrix $M$.

We then do a hyperparameter sweep over the following configurations for the architecture: 
\begin{itemize}
    \item encoder-size: small, medium, large
    \item num. input distributions: 10,30,50,75
\end{itemize}
and for the best performing architecture (large):
\begin{itemize}
    \item num. input distributions: 30,50,75
    \item learning rate: 5e-5, 7.5e-5, 1e-4
\end{itemize}
We select the model for num. input distributions 30 and learning rate 7.5e-5.

\subsubsection{GMM}
\label{sec:exp_sdd_gmm}
The GMM also has a minium-scale of $0.165$ (the trajectory domain is $[0,10]^2$).

We then do a hyperparameter sweep over the following configurations for the architecture, for both GMM-Types:
\begin{itemize}
    \item encoder-size: small, medium, large
    \item num. components: 10, 30, 50, 75
\end{itemize}
and for the winning architecture winner:
\begin{itemize}
    \item learning-rate: 5e-5, 7.5e-5, 1e-4
    \item num. components: 30, 50, 75, 150, 300, 450
\end{itemize}
Both selected a large configuration with $450$ components (learning rate 7.5e-5 for the GMM and 5e-5 for the GMM via RNN).

\subsubsection{Y-Net}
\label{sec:exp_sdd_ynet}
We train with the pretrained segmentation model, unfrozen after $15$ epochs and train for $60$, otherwise following the training recipe.

We do a hyperparameter sweep over:
\begin{itemize}
    \item learning-rate: 5e-5, 1e-4
    \item waypoint (at step): 1, 2, 3
\end{itemize}
We selected learning-rate of 1e-4 and a waypoint at 3.

\subsubsection{Sampling}
\label{sec:exp_sdd_sampling}

We sample on \pcPal by first doing ancestral sampling on the circuit to arrive at the input-distributions over the groups, so in SDD $4$ $2$-dimensional distributions over our constraints. We then sample by first sampling the simplex to sample on using prepared constrained moments. When doing the sampling on a simplex, we sample using inverse-transform sampling on the simplex using the same routines as in \gasp to integrate over the simplex. We do this by transforming first doing a coordinate change to the unit-simplex and exploiting the fact that changing the lower-bound still still results in a simplex. By doing bisection search, which can be easily parallelized over the simplices, we can sample the $4$ $2$-dimensional points in parallel and then transform back into the constraint-coordinates.

\subsubsection{Generalization To Unseen Scenes}
\label{sec:exp_sdd_ood}

In this section, we will investigate whether the predicted distribution of \pcPal transfers to unseen scenes. We provide the results (LL and sampling based metrics) in table \ref{tab:sdd_trajects_comparison_ood}.

\begin{table}[ht]
    \centering
    {
    \begin{tabular}{lccccc}
        \hline
        model & LL & ADE@1 & ADE@20 & FDE@1 & FDE@20\\
        \hline
Y-Net &	— & 26.16 & 14.46 & 37.81 & 16.76 \\
GMM & -22.68 & 34.25 & 14.70 & 56.99 & 23.12\\
GMM{\scriptsize(RNN)} & -21.31 & 15.59 & 7.55 & 27.10 & 9.10\\
\pcPal & \textbf{-21.09} & \textbf{14.93} & \textbf{6.82} & \textbf{24.59} & \textbf{7.69}\\
        \hline
    \end{tabular}
    }
    \caption{Comparison of \pcPal, GMM(RNN), GMM and Y-Net on trajectory prediction on OOD-scenes.}
    \label{tab:sdd_trajects_comparison_ood}
\end{table}

As seen in the result in \cref{tab:sdd_trajects_comparison_ood}, the constraint-handling of \pcPal transfers well do unseen scenes, with now a clearer lead in the sampling-based metrics compared to the GMM-models. We therefore arrive at an positive answer to the question whether \pcPal can generalize, as e.g. a brittleness of the distribution wrt to the changing support can not be observed.

\subsection{Full Stokes}
\label{app:full_stokes}

\subsubsection{Architecture}
\label{app:full_stokes_arch}
We do not modify the architecture by \citet{redding2013compact} and train using the original data-splits. We train using the relative MSE (to the ground-truth $S_0$), so:
$$
MSE_{polar}(y{=}(S_0,S_1,S_2,S_3),\hat{y})=MSE(y/S_0, \hat{y}/S_0)
$$
using AdamW \citep{DBLP:journals/corr/abs-1711-05101} with a learning rate of $0.001$ for $300$ epochs using early stopping. %

Using this model, we freeze the weights and enlarge the final layer to accommodate the scale for the gaussians to turn this model into a probabilistic predictor. We will only train the scale-weights in the final layer in this experiment. We train using NLL and train for $150$ epochs again using early stopping. This allows us to obtain an probabilistic predictor without interfering into the point-prediction dynamics and keep it as comparable, especially in the MAP-setting.

\subsubsection{Cone Constraints}
\label{app:full_stokes_cone_constraints}
A first observation is that the exact Full-Stokes constraints $\lVert \Vec{\vs} \rVert_2^2 \leq S_0^2$ ($S_0 \geq 0$) form a convex set, so any tangent half-space at the boundary will inevitably lead to an over-approximation of the exact feasible set.

We build the over-approximation by first discretizing the unit sphere into $163$ approximately uniformly distributed directions $\vn=(n_1,n_2,n_3)$. From these vectors, we generate the following inequality:
\begin{align}
    n_1S_1+n_2S_2+n_3S_3&\leq S_0
\end{align}
which follows from:
\begin{align}
    \lVert \Vec{\vs} \rVert_2^2 &\leq S_0^2\\
    \lVert S_1+S_2+S_3 \rVert_2^2 &\leq S_0^2\\
     \sqrt{S_1^2+S_2^2+S_3^2} &\leq S_0 && (S_0 > 0)\\
     \max_{|n|_2=1}\left(n_1S_1+n_2S_2+n_3S_3\right)&\leq S_0  && (\text{via Cauchy-Schwarz})
\end{align}
A conjunction over these inequalities, combined with the global bounds $0 \leq S_0 \leq 1$ and $\bigwedge_{i\geq1}(-1 \leq S_i \leq 1)$, form our \smtlra-constraints for the full-stokes experiment.

Since, due to noise, the raw data is not fully feasible wrt. to the exact constraints (max violation of $e\lVert S_1+S_2+S_3 \rVert_2\leq S_0$ is about $4.38e-5$), we apply a projection into the feasible set before evaluating the LL.

\subsubsection{\pcPal}
\label{app:full_stokes_pcpal}
We use a 8-piece per axis coarse grid, with a different spacing for $S_0$ to fit the data better:
For $S_0$, we use the following nonuniform knots:
$[0,\ 0.00061055,\ 0.00719695,\ 0.03047140,\ 0.08483485,\ 0.18771386,\ 0.35918497,\ 0.62172039,\ 1]$. For $(S_1,S_2,S_3)$ use $9$ equidistance knots between $-1$ and $1$. We use $K=16$ and $R=3$ for the hot-cell algorithm.

\subsubsection{Disjunctive Refinement Layer}
\label{app:full_stokes_dlr}
As the disjunctive refinement-layer \citep{DBLP:conf/iclr/StoianG25} requires a variable ordering, we evaluate the relative MSE of all possible orderings on the validation dataset and choose the best performing ordering. While the constraints must be given in CNF-form, this is already given for the Cone-Constraint.

\subsubsection{Finetuning}
\label{app:full_stokes_surface_finetuning}

We finetune on the unit-polarization dataset using only the output of the corresponding group $40$. We train for up to $300$ epochs, with an early stopping with patience of $40$ epoch and a learning rate of 1e-4 via AdamW \citep{DBLP:journals/corr/abs-1711-05101}.

\subsubsection{MAP}
\label{app:full_stokes_map}
We use PaMAP \citep{DBLP:journals/corr/abs-2602-08681} to compute the MAP for full stokes, with interior point optimization (via Adam with $16$ steps, barrier strength $0.001$ and learning rate $0.001$) as the internal optimizer. We leverage the fact that our constraints are convex, therefore we can directly construct the Polytope for the cell in $H$-form and speed up the algorithm.

\subsubsection{Surface Constraints}
\label{app:full_stokes_surface_constraints}

We construct the surface from a coarse triangular approximation of the unit sphere consisting of $20$ faces (icosahedron). We refine this coarse approximation by repeatedly splitting each triangular face into four smaller triangles and then project the newly introduced vertices back onto the unit sphere. After three such subdivisions, this yields a near-uniform approximation of the sphere with $1280$ triangular faces.

With regards to the disjunctive refinement-layer \citep{DBLP:conf/iclr/StoianG25}, as the surface constraints are essentially a DNF, the DNF to CNF conversion (in addition to replacing the equality by two inequalities) leads to a substantial blow-up and turns the application of the DLR to the surface constraints practically infeasible as the translation would result in roughly $7^{1280}$ clauses.
\end{document}